\documentclass[10pt,twocolumn,letterpaper]{article}

\usepackage[pagenumbers]{cvpr} 

\usepackage{graphicx}
\usepackage{amsmath}
\usepackage{amssymb}
\usepackage{booktabs}
\usepackage{blindtext}
\usepackage{multirow}
\usepackage[accsupp]{axessibility}
\usepackage[linesnumbered,ruled]{algorithm2e}
\usepackage[pagebackref,breaklinks,colorlinks]{hyperref}

\usepackage[capitalize]{cleveref}
\crefname{section}{Sec.}{Secs.}
\Crefname{section}{Section}{Sections}
\Crefname{table}{Table}{Tables}
\crefname{table}{Tab.}{Tabs.}

\def\cvprPaperID{82} 
\def\confName{CVPR}
\def\confYear{2023}

\begin{document}

\title{AdaMTL: Adaptive Input-dependent Inference for Efficient Multi-Task Learning}

\author{Marina Neseem\thanks{\noindent The first two authors contributed equally to this work.}, Ahmed Agiza\footnotemark[1], Sherief Reda\\
Brown University\\
Providence, RI\\
{\tt\small \{marina\_neseem, ahmed\_agiza, sherief\_reda\}@brown.edu}
}
\maketitle

\begin{abstract}
   Modern Augmented reality applications require performing multiple tasks on each input frame simultaneously. Multi-task learning (MTL) represents an effective approach where multiple tasks share an encoder to extract representative features from the input frame, followed by task-specific decoders to generate predictions for each task. Generally, the shared encoder in MTL models needs to have a large representational capacity in order to generalize well to various tasks and input data, which has a negative effect on the inference latency. In this paper, we argue that due to the large variations in the complexity of the input frames, some computations might be unnecessary for the output. Therefore, we introduce AdaMTL, an adaptive framework that learns task-aware inference policies for the MTL models in an input-dependent manner. Specifically, we attach a task-aware lightweight policy network to the shared encoder and co-train it alongside the MTL model to recognize unnecessary computations. During runtime, our task-aware policy network decides which parts of the model to activate depending on the input frame and the target computational complexity. Extensive experiments on the PASCAL dataset demonstrate that AdaMTL reduces the computational complexity by 43\% while improving the accuracy by 1.32\% compared to single-task models. Combined with SOTA MTL methodologies, AdaMTL boosts the accuracy by 7.8$\%$ while improving the efficiency by 3.1$\times$. When deployed on Vuzix M4000 smart glasses, AdaMTL reduces the inference latency and the energy consumption by up to 21.8\% and 37.5\%, respectively, compared to the static MTL model. Our code is publicly available \footnote{https://github.com/scale-lab/AdaMTL.git}.

\end{abstract}

\vspace{-10pt}
\section{Introduction}
\label{sec:intro}
\begin{figure}[t]
  \centering
   \includegraphics[trim={10 10 12 15},clip, width=0.9\linewidth]{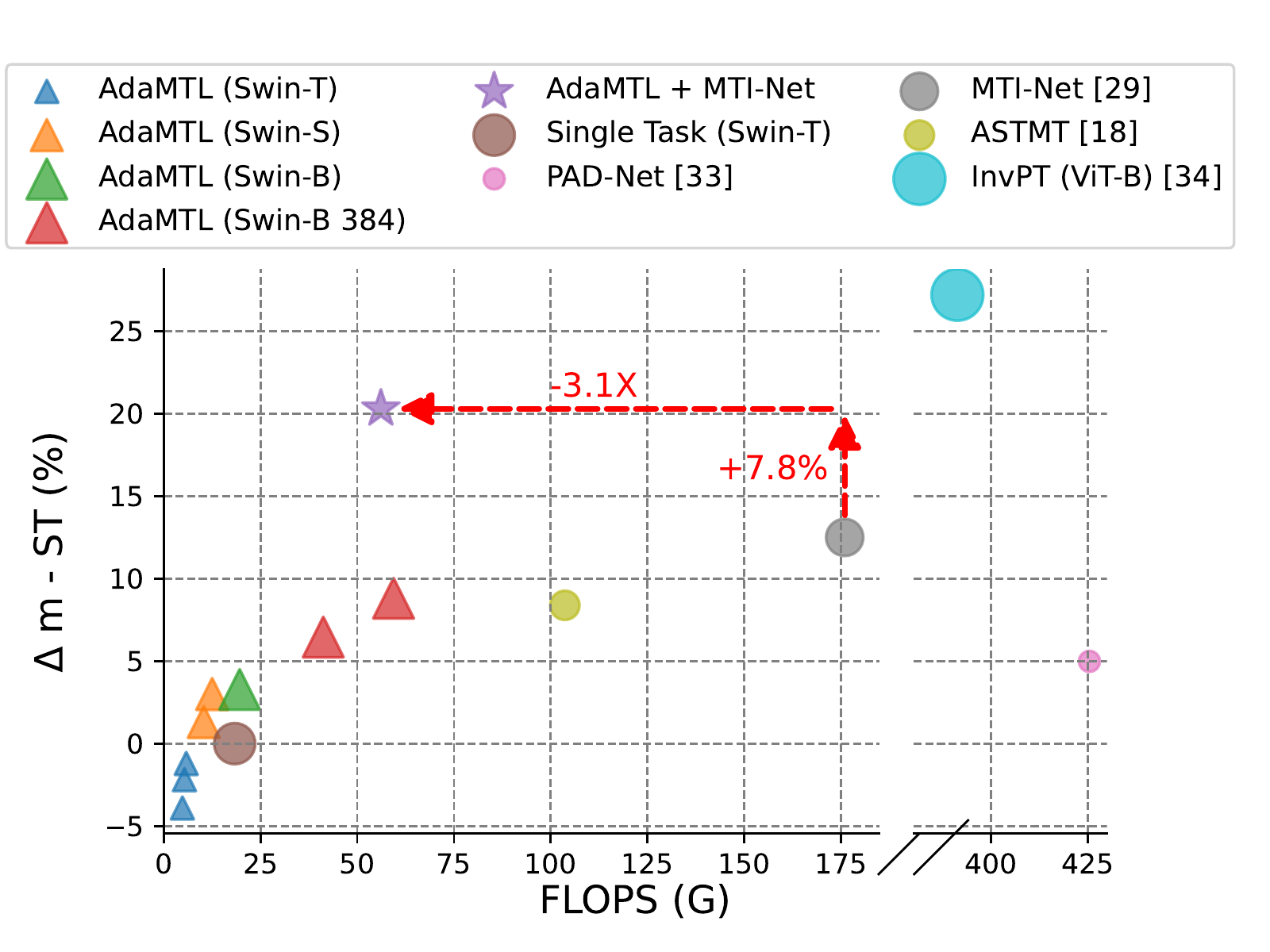}
    \vspace{-10pt}
   \caption{Accuracy-Efficiency trade-off by AdaMTL compared to SOTA MTL techniques. The x-axis shows the FLOPS, while the y-axis represents the average accuracy of the tasks compared to the single-task model.}
   \label{fig:sota}
\end{figure}

Modern computer-vision-based applications require solving multiple tasks simultaneously in order to form a complete perception of the surrounding visual environment.
For example, a simple augmented reality application might need to determine the surface normals, estimate the depths, detect an object of interest, and track it. 
Similarly, an autonomous vehicle should be able to detect both static and dynamic objects in the scene, determine their proximity and track them \cite{ishihara2021multi_auto, mohamed2021spatio}. 
Moreover, all that complex processing needs to be performed on every input frame in real-time, which is extremely challenging, especially since those applications usually run on resource-constrained devices with strict compute, memory, and energy budgets.
That is why enabling efficient computation models where these tasks can be performed simultaneously is crucial to make those applications practical. 

In recent years, Multi-task learning (MTL) has been used to learn related vision tasks simultaneously \cite{standley2020tasks, liu2019end, zamir2018taskonomy}. 
On the one hand, MTL models leverage shared representations and inter-task interactions to improve performance on each task, potentially outperforming single-task models. 
On the other hand, compared to single-task models, MTL models can reduce both the memory footprint, the energy consumption, and the latency during inference since they avoid recomputing the features in the shared layers.
Different approaches have been proposed to determine which layers should be shared across tasks and which layers should be task-specific in MTL models \cite{misra2016cross, xu2018pad, sun2020adashare}. 

One common MTL approach is to have a shared encoder that extracts the critical features from the input scene, followed by some task-specific decoders that predict the output of the corresponding task \cite{mtl_survey, vandenhende2020mti, xu2018pad}.
Generally, the shared encoder in MTL models needs to have a large representational capacity in order to generalize well to various tasks and input data from different complexities.
Vision Transformers have proven to be a powerful tool for extracting strong feature representations from the inputs, improving the performance of the downstream tasks \cite{liu2021swin, graham2021levit, deit}. 
Hence, few recent works have explored using them as a shared encoder for feature extraction in MTL models \cite{bhattacharjee2022mult, ye2022inverted}. Although these approaches achieve impressive performance outcomes, their computational complexity and memory footprint are usually huge, making them impractical for real-time processing. That's why, in this work, we focus our efforts on improving the computational complexity of transformer-based Multi-task Learning models. 

Real-world visual scenes have large variations in complexity.
For example, a scene with a single object and an open background would be easier to process than one with an occluded object and a cluttered background. 
We argue that treating all input frames equally regarding processing complexity can be wasteful. 
Therefore, we propose using an adaptive inference policy to reduce the computational complexity of MTL models based on the input complexity.
Dynamic Input-dependent inference policies have only been explored for single-task image classification problems \cite{yu2022mia, meng2022adavit}.
However, MTL models are more complex due to inter-task dependencies and complex architectures.

To this end, we propose an adaptive MTL framework that recognizes the unnecessary computations in the model depending on the input complexity. We achieve this by learning an adaptive task-aware policy network that ultimately decides on which parts of the MTL model to activate during runtime. Our contributions can be summarized as follows:

\begin{itemize}
  \setlength\itemsep{0pt}
    \item We propose an adaptive Multi-task Learning framework that optimizes Transformer-based MTL models depending on the input complexity.
    \item We introduce a task-aware policy network, and we show that it is more effective in recognizing the unnecessary computations in MTL models compared to task-agnostic policy networks.
    \item To prove that our policy network can be easily plugged to improve MTL models' efficiency, we combine AdaMTL with a SOTA MTL model \cite{vandenhende2020mti} and show that AdaMTL boosts the accuracy by 7.8\% while improving the efficiency by 3.1$\times$. 
    \item We deploy AdaMTL on the Vuzix M4000 AR glasses \cite{vuzix}, reducing the inference latency and the energy consumption by up to 21.8\% and 37.5\%, respectively, compared to the static MTL model.
\end{itemize}

The rest of the paper is organized as follows. We review the related work in Section~\ref{sec:related_work}. Then, we introduce our adaptive input-dependent MTL framework in Section~\ref{sec:methodology}. Next, we show the qualitative and quantitative analysis of our methodology as well as the ablation study in Section~\ref{sec:experiments}. Finally, we conclude in Section~\ref{sec:conclusion}.

\begin{figure*}[t]
  \centering
   \includegraphics[trim={0 20 0 0},clip, width=0.9\linewidth]{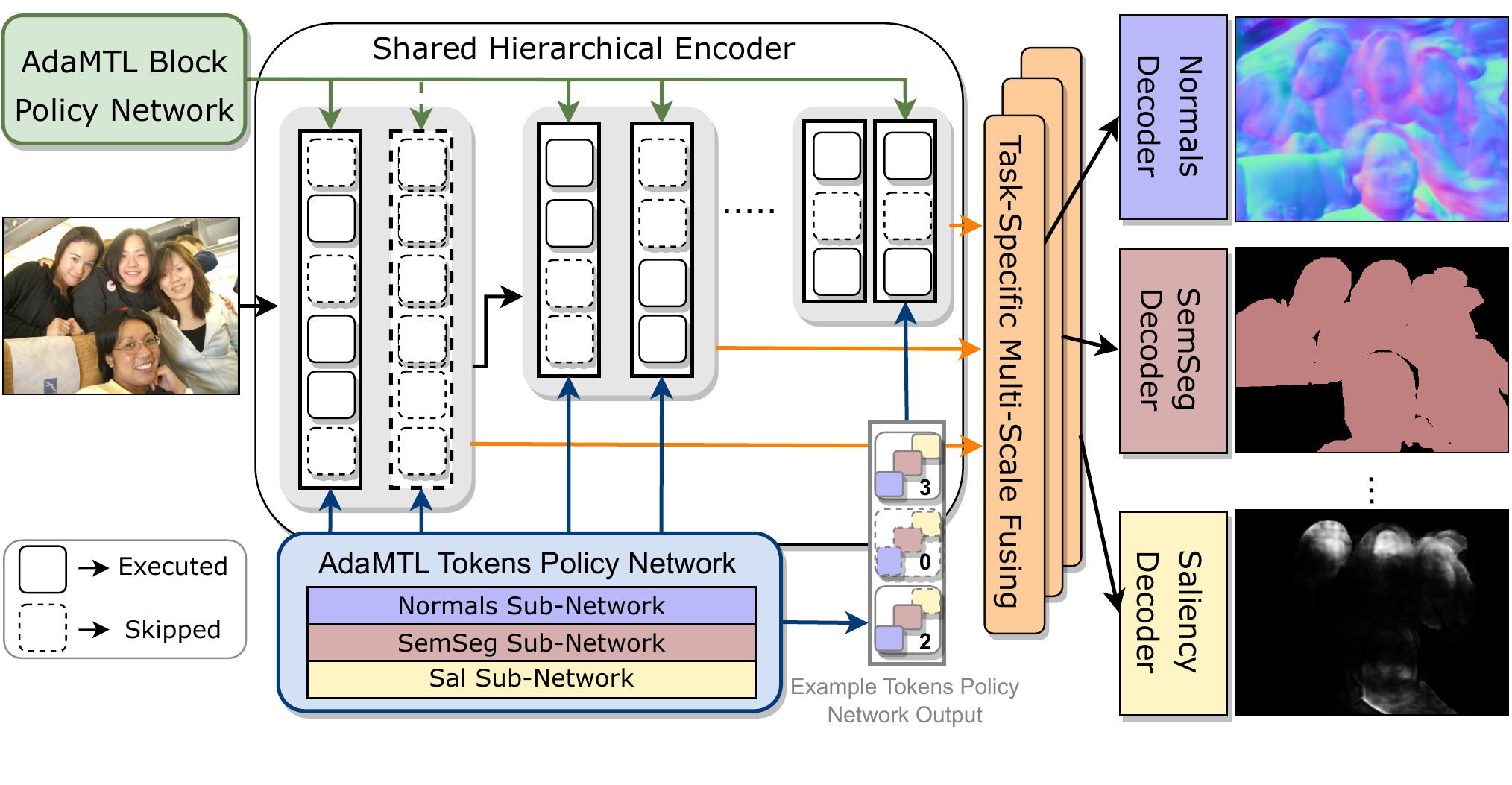}

   \caption{Overview of our proposed AdaMTL framework integrating our AdaMTL Block Policy Network and our AdaMTL Tokens Policy Network. The AdaMTL Block policy network decides on which blocks to activate during runtime. If the decision is to activate a certain block, our AdaMTL Tokens policy network runs to decide which tokens to process through that block. Our policy network achieves a task-aware behavior that improves the quality of the generated policies for multi-task learning models.}
   \label{fig:sparsemt_main_framework}
    \vspace{-15pt}
\end{figure*}
\vspace{-10pt}

\section{Background and Related Work}
\label{sec:related_work}
\textbf{Deep Multi-task Architectures}
For dense prediction tasks, deep multi-task architectures usually consist of an encoder that extracts a feature representation from the input frame, followed by task-specific decoders that generate predictions for each of the downstream tasks \cite{xu2018pad, misra2016cross, vandenhende2020mti}. 
Researchers categorize multi-task architectures based on the location of task interactions into encoder- and decoder-focused architectures \cite{mtl_survey}.
Encoder-focused architectures share the information between tasks in the encoder stage \cite{misra2016cross, mtan}, while the decoder-focused architectures exchange information between tasks in the decoding stage \cite{vandenhende2020mti, xu2018pad}.
Moreover, some approaches share the information across tasks in both encoder and decoder stages \cite{sun2020adashare, bhattacharjee2022mult}.
In this paper, our baseline MTL model follows an encoder-focused architecture where a shared encoder is used to extract visual features from the input frame, followed by task-specific decoders.
Moreover, our MTL framework can easily adopt other decoder-based task-interaction techniques, as we will show in Subsection \ref{subsec:combining_with_mtinet}.

\textbf{Vision Transformer Models}
Transformers \cite{vaswani2017attention} have achieved impressive performance improvements in various domains such as language understanding \cite{devlin2018bert}, speech recognition \cite{gulati2020conformer}, and computer vision \cite{vits, deit}. 
The self-attention modules in transformers have proved to be capable of extracting strong feature representations from the inputs, improving the performance of the downstream tasks.
To preserve local visual context, Vision Transformers (ViTs) split the input image into patches which are embedded as tokens \cite{vits}. 
Those tokens pass through successive ViT blocks. 
Each ViT block has a multi-head attention module followed by a multi-layer perceptron module to extract the global relationship among input tokens. 
Improvements have been made to ViTs enabling data-efficient training \cite{deit} as well as efficient inference \cite{graham2021levit}.
Inspired by ResNets \cite{resnets}, Swin Transformers \cite{liu2021swin} use a hierarchical ViT block architecture as well as shifted window attention to serve as a general-purpose backbone for computer vision tasks.
That is why, ViTs started replacing CNNs as a backbone for different computer vision tasks such as image classification \cite{deit,graham2021levit}, object detection \cite{object_detection}, and segmentation \cite{strudel2021segmenter}.
Moreover, several works proposed using them for multi-task learning \cite{bhattacharjee2022mult, ye2022inverted}.

\textbf{Sparsely-Activated Vision Models}
As computer vision models become increasingly complex, researchers have started sparsifying the models to prevent redundant computations. 
For example, DSelect-k \cite{hazimeh2021dselect} and M$^3$ViT \cite{m3vit} proposed Mixture-of-Experts (MoE) architectures that use trainable sparse gates to activate a subset of experts depending on the given input. 
Other methods employ early-exiting strategies to adaptively allocate computations depending on input complexity \cite{bert_loses_patience, huang2017multi}.
Other dynamic models skip redundant layers \cite{wang2018skipnet}, while others use a bottleneck layer to direct the computations in an input-dependent manner \cite{neseem2021adacon}. 
More recent approaches use lightweight policy networks to generate execution strategies based on the input complexity \cite{yu2022mia, meng2022adavit}.  
Despite the effectiveness of these approaches for image classification problems, adaptive policies have not yet been explored for MTL scenarios. 
Sparsifying MTL models is more challenging; the multi-objective nature of those models complicates the overall objective of optimizing the performance of the different tasks in addition to the policy networks. 
Therefore, there is a need to explore and develop techniques that can adaptively sparsify MTL models based on input complexity while being aware of the multi-objective nature of those models.

\section{Methodology}
\label{sec:methodology}
In this section, we propose AdaMTL \--- an adaptive end-to-end Multi-task Learning framework. 
We start by presenting our transformer-based MTL architecture in Subsection \ref{subsec:overview}. 
In Subsection \ref{subsec:policy_network}, we introduce our task-aware policy network that dynamically recognizes the unnecessary computations in the MTL model depending on the input complexity.
Finally, we explain our proposed multi-staged task-aware training recipe in Subsection \ref{subsection:training_reciepe}. 

\subsection{Overview}
\label{subsec:overview}
Figure \ref{fig:sparsemt_main_framework} illustrates an overview of our proposed adaptive end-to-end Multi-task Learning framework \-- AdaMTL. 
AdaMTL consists of two main components: a static MTL model and a lightweight policy network.
Our MTL model has three parts: a Shared Hierarchical Encoder,  a learnable task-specific multi-scale fusing layer, and a pool of task-specific decoders.
We adopt an off-the-shelf hierarchical Vision Transformer \textit{Swin} \cite{liu2021swin} as our shared encoder to extract visual features from the input frames. 
Vision Transformers (ViTs) usually split the input image into patches which are embedded as tokens. 
Those tokens pass through successive ViT blocks. 
Each ViT block has a multi-head attention module followed by a
multi-layer perceptron module to extract the global relationship among input tokens.
Our learnable multi-scale fusing layers use a residual blocks-based architecture \cite{resnets}. They are added to combine the features at different scales (i.e., receptive fields) in an informative way for every downstream task.
Finally, we use simple task-specific decoders consisting of two convolutional layers. Each decoder takes the output from the corresponding multi-scale fusing layer to generate the corresponding task predictions.

On top of our static MTL model, AdaMTL adds a lightweight policy network that runs alongside the original model to decide which parts of the model to activate, adapting to the complexity of the input frame. 
Our policy network works as a multi-grained decision maker; it consists of an \textit{AdaMTL Block policy network} and an \textit{AdaMTL Tokens policy network}. 
The \textit{AdaMTL Block policy network} decides on which blocks to activate during runtime.
If the decision is to activate a certain block, our \textit{AdaMTL Tokens policy network} runs to decide which tokens to process through that block. 
Our intuition behind AdaMTL is that not all patches in the input frame are equally informative; for example, a patch of an input frame from the background is less informative than a patch with a person for a task such as human-parts detection. 
Moreover, the receptive field at which different patches should be processed differs depending on the scale of the objects in the input frame. 
For example, an image with multiple smaller objects might benefit from being processed by the first few layers of the models where the receptive field is small. 
However, the later layers with larger receptive fields are essential for an accurate output on a zoomed-in frame with one object.
In other words, our \textit{AdaMTL Policy Network} decides which patches are needed to accurately perform the downstream tasks and the receptive field at which those patches should be processed.

\subsection{AdaMTL Policy Network}
\label{subsec:policy_network}
Our policy network works as a multi-grained decision maker; the \textit{Block Policy Network} decides on which blocks to activate, while the \textit{Tokens Policy Network} decides on which tokens (i.e., patches) to process through the activated blocks.
Our \textit{Block Policy Network} generates a learnable binary mask for each block in the shared encoder. We use this binary mask to recognize the necessary blocks needed by the MTL model in order to perform well on the downstream tasks. 
Similarly, for each block, we attach a \textit{Tokens Policy Network} that generates a learnable policy to determine the tokens that need to be processed through the activated block.
Intuitively, the policy network should learn to recognize the most informative patches in every input frame as well as the receptive field at which it needs to be processed.
Each policy network has two simple fully-connected layers, followed by a Gumbel Softmax activation to generate binary masks \cite{gumbel}.
We devise two different settings for the policy network: a \textit{task-agnostic} policy network and a \textit{task-aware} policy network.

\noindent \textbf{Task-agnostic Policy}:
In the \textit{task-agnostic} setting, the policy network is unaware of the number of downstream tasks. We achieve this by co-training the whole policy network alongside the MTL model. We mainly experiment with this setting to show the necessity of task awareness while creating an effective policy network for multi-task scenarios.

\vspace{1pt}

\noindent \textbf{Task-aware Policy}:
In the \textit{task-aware} setting, we want our policy network to capture task-specific computational needs. We achieve this by dividing the policy network into sub-networks, where each sub-network is responsible for recognizing the necessary blocks/tokens for the corresponding task.
As shown in Figure \ref{fig:sparsemt_main_framework}, the \textit{AdaMTL Tokens Policy Network} consists of a sub-network for each task (i.e., Normals Controller, Semantic Segmentation Controller, Saliency Controller, etc.). Each sub-network makes a decision that is plausible to its respective task. Finally, to get a unified policy, we make a decision to activate a token if at least one of the tasks needs to process it. The intuitive way to combine the masks would be to apply the logical ORing operation on all the generated task-specific masks. However, to make it more learnable, we combine the masks using addition and clamping. As shown in the example output from the Tokens Policy Network in Figure \ref{fig:sparsemt_main_framework}, the policy network only activates a token if at least one task-specific policy sub-network decides to activate it.

\subsection{AdaMTL Training Recipe}
\label{subsection:training_reciepe}
For our adaptive MTL framework to work effectively, we need to learn the MTL model weights as well as the execution policy (i.e., policy network’s binary masks) that achieves the target efficiency without compromising the accuracy of the various tasks in our MTL model. 
Our end-to-end training recipe consists of 3 stages:
\vspace{1pt}

\noindent \textbf{Stage 1: Static MTL Model Training}

\noindent First, we train a static MTL model. We adopt a shared encoder along with task-specific decoders to perform multi-task learning as explained in Subsection \ref{subsec:overview}.
For our shared encoder, we use the publicly-available pre-trained Swin Transformer backbones \cite{liu2021swin}. 
Then, we attach the multi-scale fusing layer as well as task-specific decoders, and we fine-tune the end-to-end MTL model to get our static baseline.
In this stage, our loss function is the weighted sum of the losses of the various downstream tasks as follows.
\vspace{-5pt}
\begin{equation}
  L_{stage1} = \sum_{i}^m \omega_{task\_i} \times L_{task\_i}
  \label{eq:stage1_loss}
\end{equation} 
\vspace{-15pt}

\noindent where $\omega_{task\_i}$ and $L_{task\_i}$ are the weight and the loss of the various tasks in the MTL model, respectively, and $m$ is the number of downstream tasks in the MTL model. We adopt the task weights used by Vandenhende \textit{et. al.} \cite{vandenhende2020mti}.

\vspace{1pt}

\noindent \textbf{Stage-2 Policy Network Initialization}

\noindent We aim to co-train both the policy network as well as the MTL model. 
Randomly initializing the policy network while co-training can lead to degrading the model accuracy since the policy network would make random decisions in the earlier epochs. 
That’s why we choose initialization weights for the policy network that activates all the blocks and the tokens. 
To achieve this, we freeze the static MTL model and pre-train our AdaMTL policy network with the following loss function:
\vspace{-8pt}
\begin{equation}
  L_{stage2} = \sum_{k}^{blocks} (1 - M_{b/k}) +  \sum_{k}^{blocks} (1 - M_{t/k}) 
  \label{eq:stage2_loss}
\vspace{-10pt}
\end{equation} 

Where $M_{b/k}$ and $M_{t/k}$ are the output masks generated by the Block Policy Network and the Tokens Policy Network attached to the Encoder block $k$, respectively. This results in an adaptive model that behaves exactly like the static model, where all blocks and tokens are activated. This acts as a good initialization point to start co-training the policy network and the MTL model.

\vspace{1pt}

\noindent \textbf{Stage 3: Policy Network/MTL Model Co-training}

\noindent In this stage, we co-train the policy network along with the MTL model. 
Our goal is to learn the binary masks that our policy network should generate in order to meet the target computational budget (i.e., the target percentage of the MTL model components to be activated) while maintaining the accuracy of the downstream tasks. 
Thus, we use a multi-objective loss as shown in Equation \ref{eq:loss_cotraining}. 
Our loss function incorporates the various task losses $L_{tasks}$ as well as the efficiency loss $L_{eff}$ multiplied by a factor $\alpha$. 
$\alpha$ represents the efficiency weight which controls the trade-off between accuracy and efficiency. In our experiments, we set $\alpha$ to unity; however, different values can be used to control the trade-off depending on the application requirements.
\vspace{-5pt}
\begin{equation}
  L_{stage3} = L_{tasks} + \alpha L_{eff}
  \label{eq:loss_cotraining}
\end{equation}

The efficiency loss incorporates the efficiency of the decisions made by the blocks as well as the tokens controller, as shown in Equation \ref{eq:loss_efficiency}. In order to minimize the number of activated blocks, we set $L_{blocks}$ as mean squared error (MSE) between the actual percentage of the activated blocks and the target percentage of activated blocks as shown in Equation \ref{eq:loss_blocks_cotraining}.
Similarly, we set $L_{tokens}$ as the MSE between the actual percentage of the activated tokens and the target percentage of activated tokens. 
However, it is common for hierarchical ViTs to have more tokens in the earlier layers compared to the later layers (i.e., the earlier layers have smaller receptive fields, thus more patches, while later layers have larger receptive fields, thus fewer patches).
This means that the number of tokens does not linearly reflect the computational complexity since layers with more tokens have smaller embedding dimensions per token, while layers with fewer tokens have larger embedding dimensions. That’s why we multiply the percentage of tokens in each layer by a weight $\omega_d$ equivalent to the embedding dimension in this layer as shown in Equation \ref{eq:loss_weighted_tokens_cotraining}.
\vspace{-10pt}
\begin{equation}
  L_{eff} = L_{blocks} + L_{tokens}
  \label{eq:loss_efficiency}
\end{equation}
\vspace{-15pt}
\begin{equation}
  L_{blocks} = MSE(\frac{B_{activ}}{B_{total}}, \frac{B_{target}}{B_{total}})
  \label{eq:loss_blocks_cotraining}
\end{equation}
\vspace{-5pt}
\begin{equation}
  L_{tokens} = MSE(\frac{\omega_d \times T_{activ}}{T_{total}}, \frac{\omega_d \times T_{target}}{T_{total}})
  \label{eq:loss_weighted_tokens_cotraining}
\end{equation}

\begin{algorithm}[t]
    \SetKwInOut{Input}{Input}
    \Input{
        $model$: Model w/ initialized policy network
        $Tasks$: Task names in MTL model \\
        $m$: Number of tasks in MTL model \\
        $epochs_{att}$: Number of epochs for ATT \\
        $L_i$: Loss for the epoch i \\
    }
    \For{$i$ in $0,\dots, epochs_{att}$}{
        $current\_task = Tasks[i \% m]$ \;
        $L_i = L_{current\_task } + \alpha L_{eff}$ \;
        \For{$task$ in $Tasks$}{
            \If{$task = current\_task$}
            {
                $\text{model.unfreeze\_decoder}(task)$ \;
                $\text{model.enable\_policy\_network}(task)$ \;
            }
            \Else
            {
                $\text{model.freeze\_decoder}(task)$ \;
                $\text{model.disable\_policy\_network}(task)$ \;
            }
         }  
        $\text{train\_one\_epoch}(model, L_i)$ \;
        
    }
    \caption{AdaMTL - Alternating Task Training (ATT)}
    \label{alg:alternating_task_training}
\end{algorithm}

\begin{table*}[t]
\small
  \centering
\renewcommand{\arraystretch}{0.90}
  \caption{Quantitative analysis of AdaMTL on PASCAL dataset. The table shows the accuracy-efficiency trade-off by our adaptive MTL model compared to the single-task model as well as static MTL models. \textit{H} and \textit{L} represent high and low computational complexity targets for AdaMTL, respectively. $\Delta$ m (ST) and $\Delta$ FLOPS (ST) show the change in the average accuracy of the tasks and percentage of FLOPS compared to the single-task model, respectively. $\downarrow$ means the lower the better, while $\uparrow$ means the higher the better. \textbf{Bolded} values represent the Pareto-frontier of the accuracy-efficiency trade-off.}
  \vspace{-10pt}
  \setlength{\tabcolsep}{3.7pt}
  \begin{tabular}{c c c | c c c c c | c c | c}
    \toprule
    \multirow{2}{*}{\textbf{Policy}} & \multirow{2}{*}{\textbf{Backbone}} & \textbf{Image} & \textbf{Sal} & \textbf{Human Parts} & \textbf{Sem Seg} & \textbf{Normals} & \textbf{$\Delta$ m (ST)} & \textbf{GFLOPS}  & \textbf{$\Delta$ FLOPS (ST)} & \textbf{Params} \\
     & & \textbf{Size} & maxF $\uparrow$ & mIoU $\uparrow$ & mIoU $\uparrow$ & mERR $\downarrow$ & (\%) $\uparrow$ & $\downarrow$ & (\%) $\downarrow$ & (M) $\downarrow$ \\
     \toprule
    Single-Task & Swin-T & 224 & 71.93 & 48.63 & 60.35 & 18.45 & 0.00 & 18.33 & 1 $\times$ & 111.42 \\
    \midrule
    Static MTL & \multirow{3}{*}{Swin-T} & 224 & 74.15 & 47.62 & 59.08 & 19.20 & \textbf{-1.20} & 5.79 & \textbf{0.32 $\times$} & 34.77 \\
    AdaMTL (H) & & 224 & 73.72 & 47.64 & 57.13 & 19.16 & \textbf{-2.18} & 5.34 & \textbf{0.29 $\times$} & 34.87 \\
    AdaMTL (L) & & 224 & 73.01 & 46.85 & 55.9 & 19.54 & \textbf{-3.86} & 4.82 & \textbf{0.26 $\times$ }& 34.87 \\

    \midrule
    Static MTL & \multirow{3}{*}{Swin-S} & 224 & 75.00 & 50.66 & 61.84 & 18.71 & +2.35 & 12.5 & 0.68 $\times$ & 67.02 \\
    AdaMTL (H) & & 224 & 75.23 & 51.22 & 61.88 & 18.79 & \textbf{+2.65} & 12.51 & \textbf{0.66 $\times$} & 67.12 \\
    AdaMTL (L) & & 224 & 74.75 & 50.07 & 59.91 & 18.61 & \textbf{+1.31} & 10.35 & \textbf{0.57 $\times$} & 67.12 \\

    \midrule
    Static MTL & \multirow{3}{*}{Swin-B} & 384 & 73.93 & 56.68 & 67.47 & 17.69 & \textbf{+8.81} & 59.39 & \textbf{3.24$\times$} & 108.66 \\
    AdaMTL (H) & & 384 & 76.55 & 56.28 & 65.04 & 17.74 & \textbf{+8.43} & 51.29 & \textbf{2.80$\times$} & 108.88 \\
    AdaMTL (L) & & 384 & 76.23 & 55.04 & 62.63 & 17.92 & \textbf{+6.46} & 41.229 & \textbf{2.25$\times$} & 108.88  \\
    
    \bottomrule
  \end{tabular}

  \label{tab:quantitative_analysis_on_pascal}
\vspace{-15pt}
\end{table*}

To make our policy network task-aware, we co-train each task-specific policy sub-network independently with the end-to-end model. 
Sequentially co-training the policy sub-networks along with the end-to-end model suffers from catastrophic forgetting (i.e., the model gets biased towards behaving well on the last task, and the performance deteriorates on the earlier tasks). 
That’s why we propose an \textit{Alternating Task Training} (ATT) where we shift the focus between the tasks every one epoch.
Algorithm \ref{alg:alternating_task_training} shows the steps of our ATT technique.
For each epoch, we choose a task to focus on, and we set the loss accordingly, as shown in lines 2 and 3, respectively.
Then, we only activate the decoder and the policy sub-network corresponding to the chosen task, as shown in lines 4-13.
We train the MTL model for one epoch using that setting. 
Then, we move on to co-training the decoders and the policy sub-networks of the other tasks. 
Finally, we perform end-to-end fine-tuning to improve the overall model accuracy. The training loss remains the same as in Equation \ref{eq:loss_cotraining}, and we unfreeze all the model’s components.
This results in an adaptive MTL model that generates task-aware inference policies depending on the complexity of the input.

\section{Experiments}
\label{sec:experiments}

\subsection{Setup}
\noindent \textbf{Dataset:} 
We evaluate our method on the PASCAL dataset \cite{pascal}. Following other papers in MTL literature \cite{ye2022inverted, vandenhende2020mti, xu2018pad}, we use the PASCAL-Context split that has annotations for various dense prediction tasks such as semantic segmentation, human part detection, surface normals estimation, and saliency distillation.
It has 4,998 images in the training split and 5,105 in the validation split.


\noindent \textbf{Implementation and Training details:} 
We implemented AdaMTL using PyTorch. 
As mentioned in Subsection \ref{subsec:overview}, we adopt the publicly available pre-trained Swin Transformer backbone \cite{liu2021swin} as our shared encoder.
To get our adaptive MTL model, we employ a three-stage training recipe as explained in Subsection \ref{subsection:training_reciepe}. 
In Stage 1, we fine-tune the Swin Transformer backbone along with our task-specific decoders for 1000 epochs. 
Then in stage 2, we freeze the MTL model and initialize the policy network by training it to activate the whole MTL model. We run this stage for another 80 epochs. 
Finally, in stage 3, we use our proposed \textit{Alternating Task Training technique} to co-train the policy network along with the MTL model for another 150 epochs, followed by fine-tuning the end-to-end AdaMTL model for another 150 epochs.
Our method does not only increase the model efficiency during inference, but it also reduces the carbon emission since we avoid retraining complex ViTs from scratch by reusing off-the-shelf pre-trained backbones; AdaMTL needs only 1 V100 GPU for around 24-48 hours (i.e., depending on the used backbone) in order to run our end-to-end training recipe.

\subsection{Quantitative Analysis}
\noindent \textbf{Accuracy-Efficiency Trade-off:}
 We apply AdaMTL on three SOTA ViTs from the Swin Transformer family \cite{liu2021swin}. We include \textit{Swin-Tiny}, \textit{Swin-Small}, and \textit{Swin-Base}, representing three different scales of ViTs in terms of computational complexity.
We include two different computational complexity targets: \textit{H} and \textit{L}. \textit{H} represents a higher computational budget where the target percentage of activated tokens and blocks are 60\% and 90\%, respectively. \textit{L} represents a lower computational budget where both the target percentage of activated tokens and blocks are set to 50\%. In both settings, the accuracy-efficiency trade-off weight (i.e., $\alpha$ in Equation \ref{eq:loss_cotraining}) is set to unity.
To evaluate the accuracy-efficiency trade-off by AdaMTL, we compare it to two baselines: (1) Single-Task models and (2) Static MTL models (i.e., our base MTL model before attaching our task-aware policy network).
Table \ref{tab:quantitative_analysis_on_pascal} shows the accuracy, the computational complexity (i.e., FLOPS) as well as the model size (i.e., Params) of AdaMTL compared to the single-task model and the static MTL models.
$\Delta$ m (ST) and $\Delta$ FLOPS (ST) show the change in the average accuracy of the tasks and percentage of FLOPS compared to the single-task model, respectively.
Results show how our method enhances the accuracy-efficiency trade-off for MTL models. 
For example, by applying AdaMTL to \textit{Swin-S} backbone, we can get an MTL model with 43\% less FLOPS and 1.31\% more accuracy compared to the single-task model.
Similarly, by applying AdaMTL to \textit{Swin-T} backbone, we can get an MTL model with 71\% less FLOPS and only 2.18\% drop in accuracy compared to the single-task model.
Therefore, given any target computational complexity, AdaMTL can meet it while potentially improving the accuracy.

\begin{table}[t]
  \centering
  \setlength{\tabcolsep}{2.2pt}
\renewcommand{\arraystretch}{0.85}
  \caption{Comparison with SOTA MTL models.}
  \vspace{-10pt}
  \small
  \begin{tabular}{c c c c c}
    \toprule
    \multirow{2}{*}{\textbf{Method}} & \multirow{2}{*}{\textbf{Backbone}} & \textbf{$\Delta$ m} (ST) & \textbf{$\Delta$ FLOPS (ST)} &\\
    & & $\uparrow$ (\%) & $\downarrow$ (\%) \\
    \midrule
    PAD-Net \cite{xu2018pad} & HRNet-18  & +4.98 & 23.21$\times$ \\
    AdaMTL (Ours) & Swin-B & \textbf{+6.46} & \textbf{2.25 $\times$} \\
    \midrule
    ASTMT \cite{astmt} & R26-DLv3 & +8.38 & 5.66 $\times$ \\
    AdaMTL (Ours) & Swin-B & \textbf{+8.81} & \textbf{3.24 $\times$} \\
    \midrule
    MTI-Net \cite{vandenhende2020mti} & ResNet-50 & +13.49 & 9.6 $\times$ \\
    InvPT \cite{ye2022inverted} & ViT-B & +27.20 & 21.35 $\times$ \\
    \bottomrule
  \end{tabular}
  \label{tab:comparison_to_sota}
\end{table}

\begin{figure*}
  \centering
  \begin{subfigure}{0.6\linewidth}
  \begin{subfigure}{0.24\linewidth}
    \includegraphics[height=\linewidth, width=\linewidth, keepaspectratio]{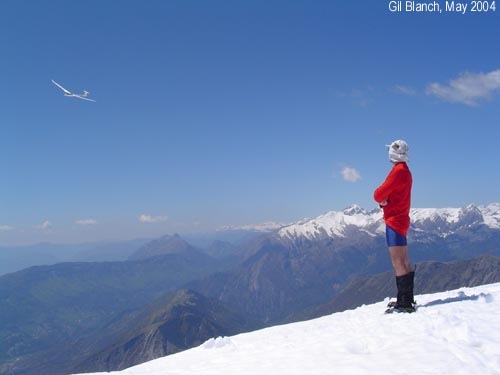} 
    \includegraphics[height=\linewidth, width=\linewidth, keepaspectratio]{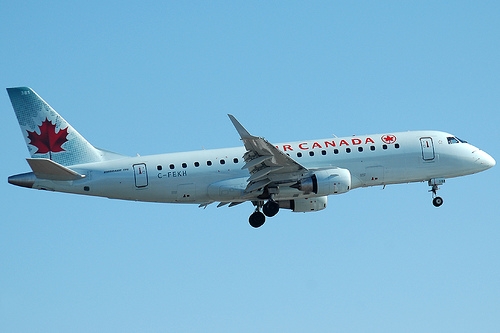}
    \caption{55\% FLOPS}
    \label{fig:qualitative_55_flops}
  \end{subfigure} 
  \hfill
  \begin{subfigure}{0.24\linewidth}
    \includegraphics[height=\linewidth, width=\linewidth, keepaspectratio]{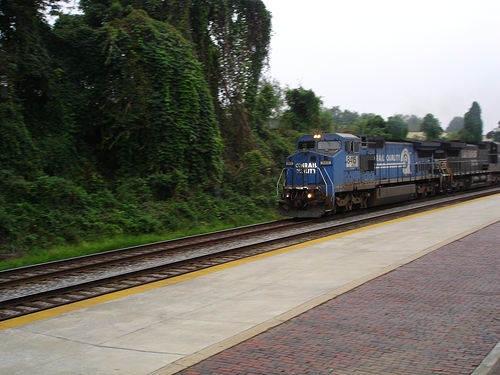} 
    \includegraphics[height=\linewidth, width=\linewidth, keepaspectratio]{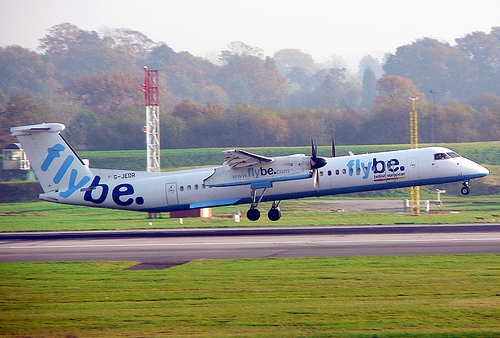}
    \caption{65\% FLOPS}
    \label{fig:qualitative_65_flops}
  \end{subfigure} 
  \hfill
  \begin{subfigure}{0.24\linewidth}
    \includegraphics[height=\linewidth, width=\linewidth, keepaspectratio]{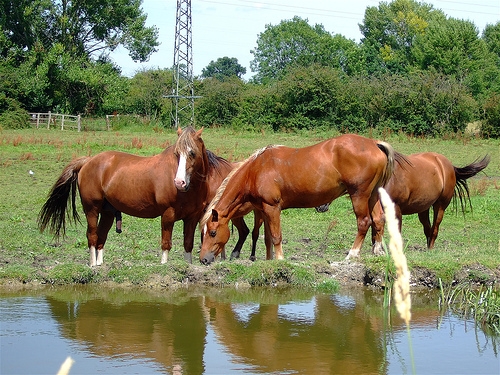} 
    \includegraphics[trim={0 0 0 40},clip, height=\linewidth, width=\linewidth, keepaspectratio]{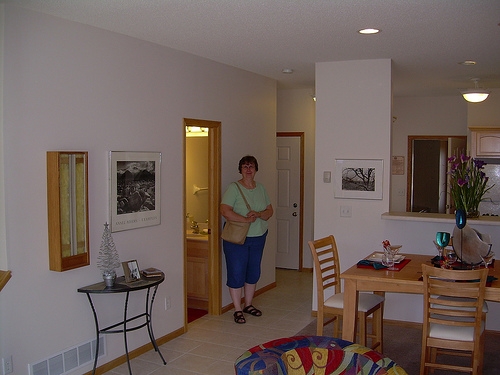}
    \caption{75\% FLOPS}
    \label{fig:qualitative_75_flops}
  \end{subfigure} 
  \hfill
  \begin{subfigure}{0.24\linewidth}
    \includegraphics[height=\linewidth, width=\linewidth, keepaspectratio]{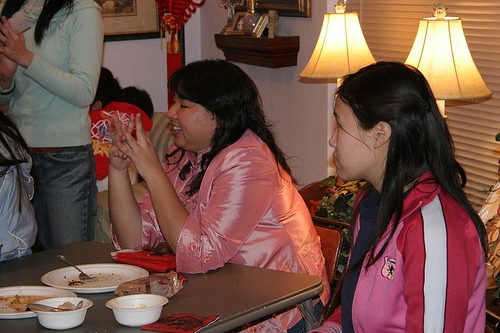} 
    \includegraphics[height=\linewidth, width=\linewidth, keepaspectratio]{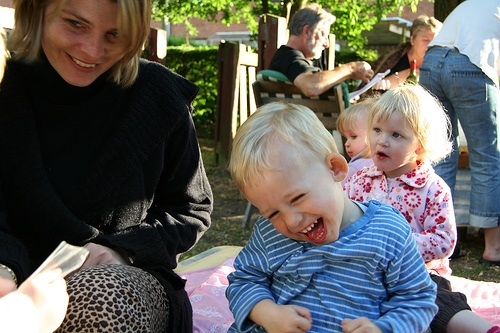}
    \caption{85\% FLOPS}
    \label{fig:qualitative_85_flops}
  \end{subfigure} 

  \end{subfigure} 
  \hfill
  \begin{subfigure}{0.38\linewidth}
   \includegraphics[width=\linewidth]{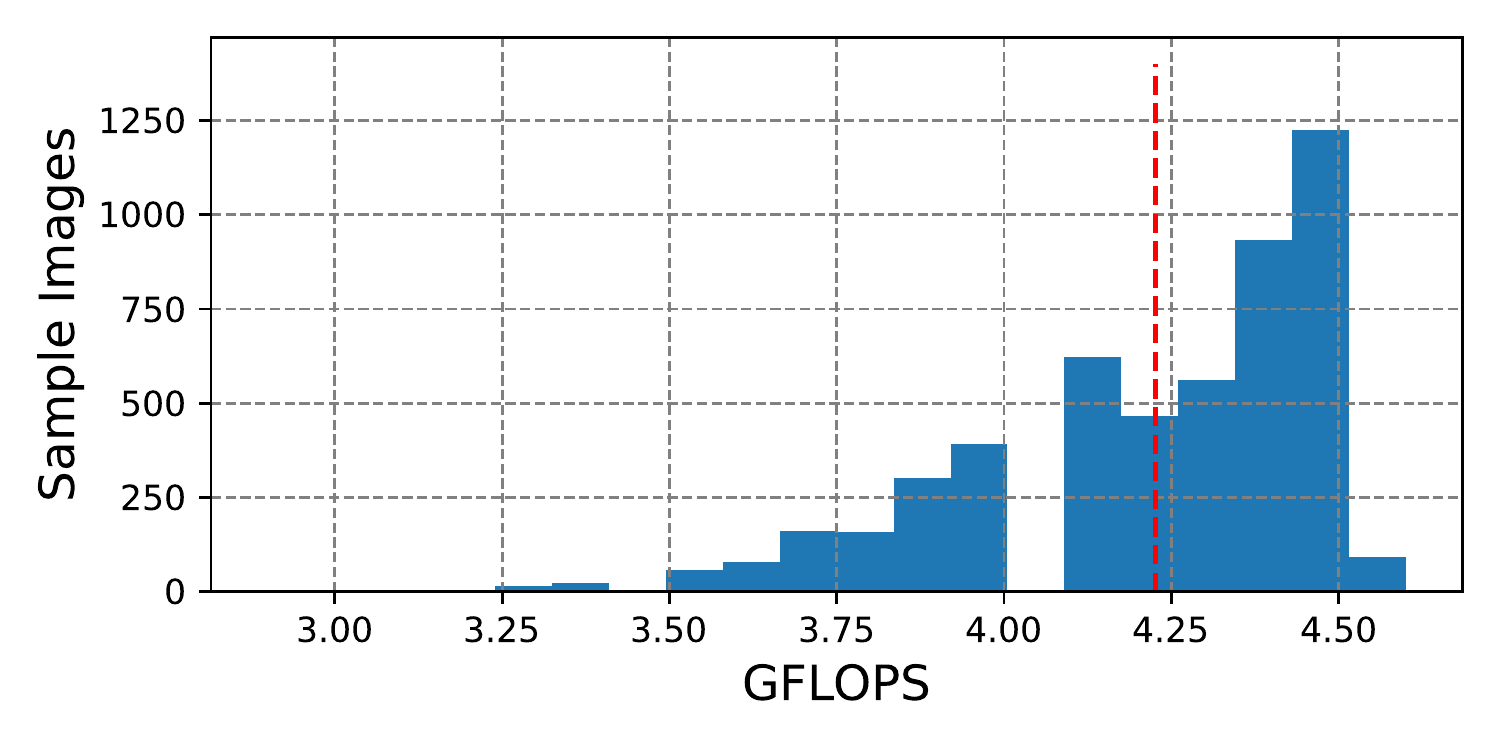}
   \caption{GFLOPs Histogram of AdaMTL on PASCAL dataset}
    \label{fig:qualitative_histogram}
  \end{subfigure}
  \caption{Qualitative insights on the computational budget allocated by AdaMTL to process input frames of different complexity. We can notice that AdaMTL assigns fewer computations for simpler scenes, as in (a) and (b), while it justly assigns more computations to process more complex and cluttered scenes as in (c) and (d).}
  \label{fig:qualitative_insights}
  \vspace{-15pt}
\end{figure*}

\noindent \textbf{Comparison to SOTA MTL models:}
We also compare the accuracy-efficiency trade-off of AdaMTL to four SOTA MTL models that vary in computational complexity and accuracy.
Figure \ref{fig:sota} shows the accuracy and the computational complexity of AdaMTL compared to those of \textit{PAD-Net} \cite{xu2018pad}, \textit{ASTMT} \cite{astmt}, \textit{MTI-Net} \cite{vandenhende2020mti}, and \textit{InvPT} \cite{ye2022inverted}.
We can notice that AdaMTL dominates both \textit{PAD-Net} and \textit{ASTMT}. Moreover, AdaMTL achieves a more efficient trade-off compared to \textit{MTI-Net} and \textit{InvPT}. 
As shown in Table \ref{tab:comparison_to_sota}, AdaMTL has 3$\times$ less FLOPS than MTI-Net and 7$\times$ less FLOPS than \textit{InvPT}. It is important to note that our effort in AdaMTL is directed specifically toward enabling efficient Multi-Task learning. That is why we argue that while the performance of \textit{MTI-Net} and \textit{InvPT} is impressive, their demanding computational complexity might not be suitable for real-time processing on resource-constrained devices.

\begin{table}[t]
  \centering
  \small
   \setlength{\tabcolsep}{1.7pt}
\renewcommand{\arraystretch}{0.85}
  \caption{Combining AdaMTL with SOTA MTL components.}
  \vspace{-10pt}
  \begin{tabular}{c c | c c c }
    \toprule
    \multirow{2}{*}{\textbf{Method}} & \multirow{2}{*}{\textbf{Backbone}} & \textbf{$\Delta$ m} & \textbf{$\Delta$ FLOPS} & \textbf{Params} \\
    & & \% & $\downarrow$ & (M) \\
    \midrule
    Single-Task & Swin-T & +0.00 & 18.33 G & 111.42 \\
    AdaMTL & Swin-B & +8.81 & 1.1$\times$ & 108.66 \\
    MTI-Net \cite{vandenhende2020mti} & ResNet-50 & +13.49 & 9.6$\times$ & 91.00 \\
    AdaMTL + MTI-Net & Swin-B & \textbf{+20.29} & \textbf{3.1$\times$} & 101.14 \\    
    \bottomrule
  \end{tabular}
\vspace{+3pt}
  \label{tab:combining_with_mti_net}
\end{table}

\subsection{Combining with SOTA MTL components}
\label{subsec:combining_with_mtinet}
Our adaptive MTL framework can easily adopt other SOTA MTL components to further enhance the accuracy-efficiency trade-off. 
In this section, we show a case study where we integrate AdaMTL with the SOTA MTL concepts in \textit{MTI-Net} \cite{vandenhende2020mti} in order to improve both its efficiency and accuracy. 
\textit{MTI-Net} has two main modules: (1) A multi-scale multi-modal distillation unit to model task interactions at different scales and (2) A feature propagation module that propagates distilled task information from lower to higher scales.
In this experiment, we attach those two modules between the shared hierarchical encoder and the task-specific decoders in our AdaMTL framework. 
Following the exact same training recipe introduced in Subsection \ref{subsection:training_reciepe}, the results in Table \ref{tab:combining_with_mti_net} show that AdaMTL can be integrated with other MTL modules from \textit{MTI-Net} to boost \textit{MTI-Net}'s accuracy by 7.8$\%$ while improving its efficiency by 3.1$\times$.


\begin{figure}
  \centering
  \begin{subfigure}{0.24\linewidth}
    \includegraphics[height=\linewidth, width=\linewidth]{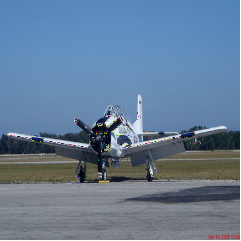} 
    \includegraphics[height=\linewidth, width=\linewidth]{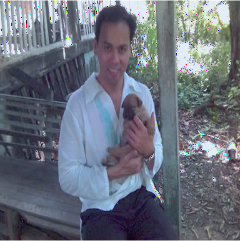}
    \caption{Input Frame}
    \label{fig:input_samples_masks}
  \end{subfigure} 
  \hfill
  \begin{subfigure}{0.24\linewidth}
    \includegraphics[height=\linewidth, width=\linewidth]{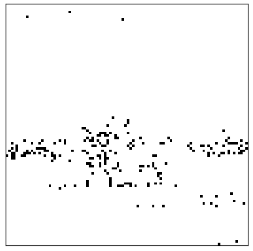} 
    \includegraphics[height=\linewidth, width=\linewidth]{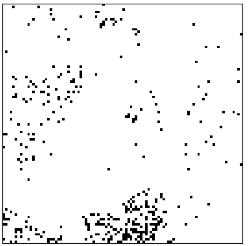}
    \caption{Layer 1}
    \label{fig:layer1_masks}
  \end{subfigure} 
  \hfill
  \begin{subfigure}{0.24\linewidth}
    \includegraphics[height=\linewidth, width=\linewidth]{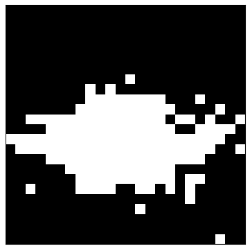} 
    \includegraphics[height=\linewidth, width=\linewidth]{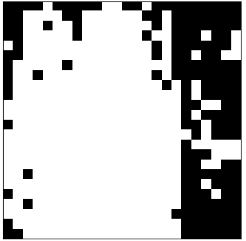}
    \caption{Layers 12-15}
    \label{fig:mid_layers_masks}
  \end{subfigure} 
  \hfill
  \begin{subfigure}{0.24\linewidth}
    \includegraphics[height=\linewidth, width=\linewidth]{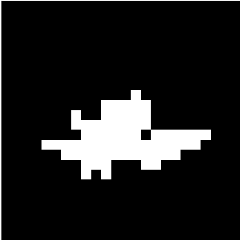} 
    \includegraphics[height=\linewidth, width=\linewidth]{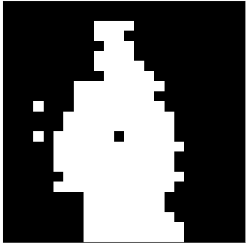}
    \caption{Layer 18}
    \label{fig:final_layers_masks}
  \end{subfigure} 
  \caption{Sample of the generated masks by AdaMTL tokens policy network. The white areas in (b)-(d) represent activated tokens.}
  \vspace{+3pt}
  \label{fig:generated_masks_by_policy_network}

\end{figure}

\subsection{Qualitative Analysis}
Figure \ref{fig:qualitative_insights} shows some insights about the allocated amount of computations by AdaMTL for input frames of different visual complexity. Figures \ref{fig:qualitative_55_flops}, \ref{fig:qualitative_65_flops}, \ref{fig:qualitative_75_flops}, and \ref{fig:qualitative_85_flops} shows examples where AdaMTL allocated 55\%, 65\%, 75\%, and 85\% of the static model FLOPS respectively. We can see that AdaMTL allocates fewer computations to simple frames with fewer objects as compared to complex scenes with multiple objects and cluttered backgrounds. Figure \ref{fig:qualitative_histogram} shows a histogram of AdaMTL's computational complexity per example for all the images in PASCAL validation set. The red line represents the average number of FLOPS needed to process all the images.
We can notice that AdaMTL adapts to the large variation in the computational complexity requirement by various images in the dataset.

To gain more insights into the decisions made by our policy network, we visualize a sample of the generated tokens masks in Figure \ref{fig:generated_masks_by_policy_network}. 
Column \ref{fig:input_samples_masks} represents the input frames, while columns \ref{fig:layer1_masks}, \ref{fig:mid_layers_masks}, and \ref{fig:final_layers_masks} represent the generated tokens masks by our policy network at different layers, such that the white areas represent activated tokens. 
We can notice that the policy network tends to activate more tokens in the earlier layers to understand the global features of the input frame.
Then, it narrows down its scope in the later layers, focusing on the most informative patches (i.e., patches with the main objects) in the input frame.

\subsection{Deployment on Vuzix M4000 AR glasses}
We compile our AdaMTL model using PyTorch for Android \cite{pytorch}, and we deploy it on the Vuzix M4000 AR glasses \cite{vuzix}. The Vuzix glasses have an 8 Core 2.52Ghz Qualcomm XR1 board with 6GB RAM. It operates using Android 11.0.  
Using the \textit{Battery Historian} tool \cite{historian} to profile the energy consumption on the device, we record the average latency and energy consumption across random samples from the PASCAL validation dataset. Table \ref{tab:vuzix_numbers} shows that AdaMTL reduces the inference latency and the energy consumption by up to 21.8\% and 37.5\%, respectively, compared to its corresponding static MTL model.

\begin{table}[t]
  \centering
  \setlength{\tabcolsep}{5pt}
\renewcommand{\arraystretch}{0.85}
  \caption{Performance Analysis on Vuzix Augmented Reality Glasses \cite{vuzix}. We analyze the percentage of the inference latency and the energy consumption by our AdaMTL model compared to its corresponding static model.}
  \vspace{-5pt}
  \small
  \begin{tabular}{c c | c c}
    \toprule
    \multirow{2}{*}{Method} & \multirow{2}{*}{Backbone} & Inference  & Energy \\
    & & Latency (\%) & Consumption (\%) \\
    \midrule
    AdaMTL & Swin-T &  -20.6\% & -35.3\% \\
    AdaMTL & Swin-S & -21.8\% & -37.5\% \\
    \bottomrule
  \end{tabular}
  \label{tab:vuzix_numbers}
\end{table}

\begin{table}[t]
  \centering
  \vspace{-10pt}
  \caption{Comparison between the quality of our task-aware policy, the task-agnostic, and the random execution policy.}
  \renewcommand{\arraystretch}{0.85}
  \vspace{-5pt}
  \small
  \begin{tabular}{c | c | c c}
    \toprule
    \textbf{Policy} & \textbf{Backbone} & \textbf{$\Delta$ m} (ST)  & \textbf{$\Delta$ FLOPS} (ST) \\
    & & (\%) $\uparrow$& $\downarrow$ \\

    \toprule
    Random & \multirow{4}{*}{Swin-T} & -34.53 & 0.24$\times$ \\
    Random$+$ & & -8.64 & 0.24$\times$ \\
    Task-Agnostic & & -5.68 & 0.24$\times$ \\
    Task-Aware & & \textbf{-3.86} & \textbf{0.26$\times$} \\
    \midrule
    Random & \multirow{4}{*}{Swin-B} & -34.86 & 0.82$\times$ \\
    Random$+$ & & -8.57 & 0.82$\times$ \\
    Task-Agnostic & & +0.77 & 0.84$\times$ \\
    Task-Aware & & \textbf{+1.12} & \textbf{0.83$\times$} \\
    \bottomrule
  \end{tabular}
  \vspace{+5pt}
  \label{tab:abalation_study_learnt_policy_quallity}
\end{table}

\subsection{Ablation Study}
\noindent \textbf{Quality of the learnt inference policies}:
To analyze the quality of the learned inference policies by our task-aware policy network. Table \ref{tab:abalation_study_learnt_policy_quallity} compares the accuracy-efficiency trade-off achieved by our task-aware policy network (i.e., referred to as Task-Aware) to three other baselines: (1) Random where we activate random blocks and tokens from the static MTL model, (2) Random$+$ where we again activate random blocks and tokens from the static MTL model but after performing our adaptive training pipeline, and (3) Task-Agnostic policy network explained in Subsection \ref{subsec:policy_network}.
The results in Table \ref{tab:abalation_study_learnt_policy_quallity} 
 show that the policies learned by our task-aware policy network outperform the other baselines. 
 We can also notice that performing adaptive training gives the MTL model robustness towards sparsification (i.e., Using a random policy on the static model reduced the accuracy by ~34\%, while it only reduced the accuracy by 8\% when applied to the adaptively trained MTL model). 

\noindent \textbf{Analysis of the adaptation along each component of our policy network}
To understand the contribution of both components of our policy network (i.e., blocks policy network and tokens policy network) to the accuracy-efficiency trade-off of AdaMTL, we compare the results from four different settings: (1) the static model where neither component of the policy network is activated, (2) AdaMTL while activating the block policy network only, (3) AdaMTL while activating the tokens policy network only, and (4) AdaMTL where both policy networks are activated. Table \ref{tab:dimensions_contribution} shows that enabling both components enhances the accuracy-efficiency trade-off of AdaMTL.

\begin{figure}[t]
  \centering
  \begin{subfigure}{0.45\linewidth}
    \includegraphics[width=\linewidth]{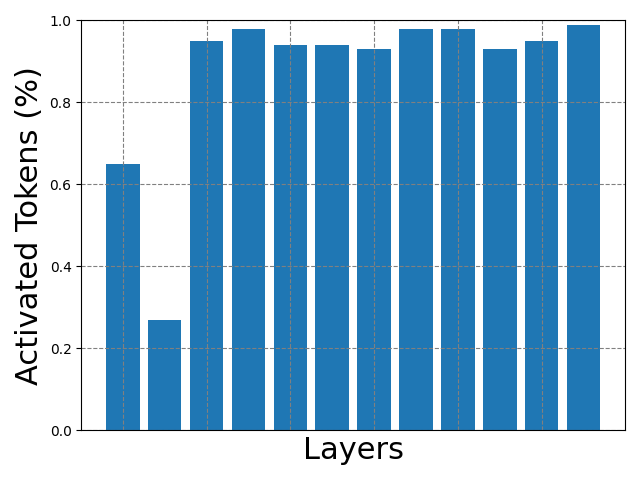}
    \caption{Without weighted tokens}
  \end{subfigure}
  \hfill
  \begin{subfigure}{0.45\linewidth}
    \includegraphics[width=\linewidth]{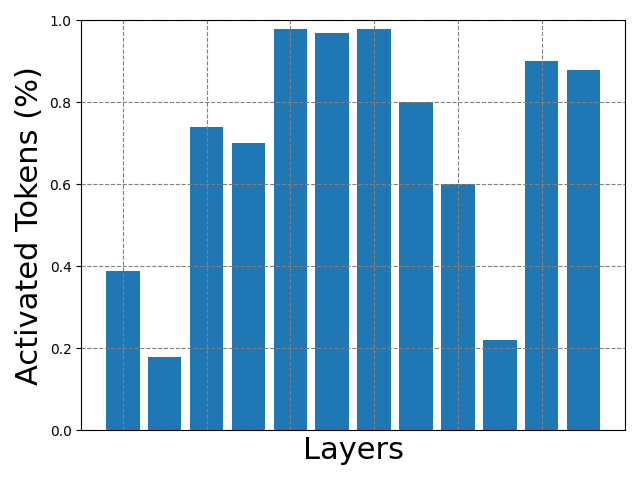}
    \caption{With weighted tokens}
  \end{subfigure}
  \caption{Loss function ablation: The figure illustrates the average percentage of activated tokens over the 12 blocks in Swin-T using (a) non-weighted tokens loss and (b) weighted tokens loss.}
  \label{fig:tokens_loss_weight_analysis}
  \vspace{+5pt}
\end{figure}

\noindent \textbf{Analysis of the behavior of our loss function}
To analyze the importance of the weight factor $\omega_d$ in Equation \ref{eq:loss_weighted_tokens_cotraining} of our loss function, we visualize the average percentage of activated tokens across the blocks of Swin-T encoder with and without $\omega_d$ in Figure \ref{fig:tokens_loss_weight_analysis}. We can notice that adding the $\omega_d$ factor to the loss function leads to a more distributed FLOPS reduction across different blocks, which is essential for the effectiveness of our adaptive MTL framework.


\begin{table}[t]
  \centering
\renewcommand{\arraystretch}{0.85}
  \caption{The contribution of different adaptive dimensions to AdaMTL. $\Delta$ m is measured relative to the static MTL model.}
  \vspace{-5pt}
  \small
  \begin{tabular}{c c | c c}
    \toprule
    Adaptive & Adaptive & $\Delta$ m & FLOPS \\
    Blocks & Tokens & \% & (G) \\
    \midrule
    $\times$ & $\times$ & -0.00 & 5.8 \\
    $\checkmark$ & $\times$ & -1.66 & 5.23 \\
    $\times$ & $\checkmark$ & -1.46 & 5.08 \\
    $\checkmark$ & $\checkmark$ & \textbf{-0.85} & 5.37 \\
    \bottomrule
  \end{tabular}

  \label{tab:dimensions_contribution}
  \vspace{-12pt}
\end{table}

\section{Conclusion}
\label{sec:conclusion}
In this paper, we propose AdaMTL \-- an adaptive framework
that learns task-aware inference policies for the MTL models in an input-dependent manner. 
We achieve this by co-training a lightweight policy network along with our MTL model. During runtime, our policy network recognizes the unnecessary computations and dynamically chooses an execution strategy depending on the input complexity and the target computational budget. Our experiments on PASCAL dataset demonstrate that AdaMTL reduces the computational complexity by 43\% while improving the accuracy
by 1.32\% compared to the single task models. Combined with SOTA MTL components, AdaMTL boosts the accuracy by 7.8\% while improving the SOTA MTL model efficiency by 3.1$\times$.
Finally, we deployed AdaMTL on Vuzix M4000 AR glasses showing up to 21.8\% and 37.5\% reduction in inference latency and energy consumption, respectively, compared to the static MTL model.

\noindent \textbf{Acknowledgements:} This work is partially supported by NSF grant 1814920 and DoD ARO grant W911NF-19-1-0484.
{\small
\bibliographystyle{ieee_fullname}
\bibliography{egbib}

\begin{thebibliography}{10}\itemsep=-1pt

\bibitem{bhattacharjee2022mult}
Deblina Bhattacharjee, Tong Zhang, Sabine S{\"u}sstrunk, and Mathieu Salzmann.
\newblock Mult: an end-to-end multitask learning transformer.
\newblock In {\em Proceedings of the IEEE/CVF Conference on Computer Vision and
  Pattern Recognition}, pages 12031--12041, 2022.

\bibitem{object_detection}
Nicolas Carion, Francisco Massa, Gabriel Synnaeve, Nicolas Usunier, Alexander
  Kirillov, and Sergey Zagoruyko.
\newblock End-to-end object detection with transformers.
\newblock In {\em Computer Vision--ECCV 2020: 16th European Conference,
  Glasgow, UK, August 23--28, 2020, Proceedings, Part I 16}, pages 213--229.
  Springer, 2020.

\bibitem{devlin2018bert}
Jacob Devlin, Ming-Wei Chang, Kenton Lee, and Kristina Toutanova.
\newblock Bert: Pre-training of deep bidirectional transformers for language
  understanding.
\newblock {\em arXiv preprint arXiv:1810.04805}, 2018.

\bibitem{vits}
Alexey Dosovitskiy, Lucas Beyer, Alexander Kolesnikov, Dirk Weissenborn,
  Xiaohua Zhai, Thomas Unterthiner, Mostafa Dehghani, Matthias Minderer, Georg
  Heigold, Sylvain Gelly, et~al.
\newblock An image is worth 16x16 words: Transformers for image recognition at
  scale.
\newblock {\em arXiv preprint arXiv:2010.11929}, 2020.

\bibitem{pascal}
Mark Everingham, Luc Van~Gool, Christopher Williams, John Winn, and Andrew
  Zisserman.
\newblock The pascal visual object classes (voc) challenge.
\newblock {\em International Journal of Computer Vision}, 88:303--338, 06 2010.

\bibitem{graham2021levit}
Benjamin Graham, Alaaeldin El-Nouby, Hugo Touvron, Pierre Stock, Armand Joulin,
  Herv{\'e} J{\'e}gou, and Matthijs Douze.
\newblock Levit: a vision transformer in convnet's clothing for faster
  inference.
\newblock In {\em Proceedings of the IEEE/CVF international conference on
  computer vision}, pages 12259--12269, 2021.

\bibitem{gulati2020conformer}
Anmol Gulati, James Qin, Chung-Cheng Chiu, Niki Parmar, Yu Zhang, Jiahui Yu,
  Wei Han, Shibo Wang, Zhengdong Zhang, Yonghui Wu, et~al.
\newblock Conformer: Convolution-augmented transformer for speech recognition.
\newblock {\em arXiv preprint arXiv:2005.08100}, 2020.

\bibitem{hazimeh2021dselect}
Hussein Hazimeh, Zhe Zhao, Aakanksha Chowdhery, Maheswaran Sathiamoorthy, Yihua
  Chen, Rahul Mazumder, Lichan Hong, and Ed Chi.
\newblock Dselect-k: Differentiable selection in the mixture of experts with
  applications to multi-task learning.
\newblock {\em Advances in Neural Information Processing Systems},
  34:29335--29347, 2021.

\bibitem{resnets}
Kaiming He, Xiangyu Zhang, Shaoqing Ren, and Jian Sun.
\newblock Deep residual learning for image recognition.
\newblock In {\em Proceedings of the IEEE conference on computer vision and
  pattern recognition}, pages 770--778, 2016.

\bibitem{historian}
Battery historian, August 2016.
\newblock \url{https://github.com/google/battery-historian}.

\bibitem{huang2017multi}
Gao Huang, Danlu Chen, Tianhong Li, Felix Wu, Laurens Van Der~Maaten, and
  Kilian~Q Weinberger.
\newblock Multi-scale dense networks for resource efficient image
  classification.
\newblock {\em arXiv preprint arXiv:1703.09844}, 2017.

\bibitem{ishihara2021multi_auto}
Keishi Ishihara, Anssi Kanervisto, Jun Miura, and Ville Hautamaki.
\newblock Multi-task learning with attention for end-to-end autonomous driving.
\newblock In {\em Proceedings of the IEEE/CVF Conference on Computer Vision and
  Pattern Recognition}, pages 2902--2911, 2021.

\bibitem{gumbel}
Eric Jang, Shixiang Gu, and Ben Poole.
\newblock Categorical reparameterization with gumbel-softmax.
\newblock {\em arXiv preprint arXiv:1611.01144}, 2016.

\bibitem{m3vit}
Hanxue Liang, Zhiwen Fan, Rishov Sarkar, Ziyu Jiang, Tianlong Chen, Kai Zou, Yu
  Cheng, Cong Hao, and Zhangyang Wang.
\newblock M\textsuperscript{3}vit: Mixture-of-experts vision transformer for
  efficient multi-task learning with model-accelerator co-design.
\newblock {\em arXiv preprint arXiv:2210.14793}, 2022.

\bibitem{liu2019end}
Shikun Liu, Edward Johns, and Andrew~J Davison.
\newblock End-to-end multi-task learning with attention.
\newblock In {\em Proceedings of the IEEE/CVF conference on computer vision and
  pattern recognition}, pages 1871--1880, 2019.

\bibitem{mtan}
Shikun Liu, Edward Johns, and Andrew~J Davison.
\newblock End-to-end multi-task learning with attention.
\newblock In {\em Proceedings of the IEEE/CVF conference on computer vision and
  pattern recognition}, pages 1871--1880, 2019.

\bibitem{liu2021swin}
Ze Liu, Yutong Lin, Yue Cao, Han Hu, Yixuan Wei, Zheng Zhang, Stephen Lin, and
  Baining Guo.
\newblock Swin transformer: Hierarchical vision transformer using shifted
  windows.
\newblock In {\em Proceedings of the IEEE/CVF international conference on
  computer vision}, pages 10012--10022, 2021.

\bibitem{astmt}
Kevis-Kokitsi Maninis, Ilija Radosavovic, and Iasonas Kokkinos.
\newblock Attentive single-tasking of multiple tasks.
\newblock In {\em Proceedings of the IEEE/CVF Conference on Computer Vision and
  Pattern Recognition}, pages 1851--1860, 2019.

\bibitem{meng2022adavit}
Lingchen Meng, Hengduo Li, Bor-Chun Chen, Shiyi Lan, Zuxuan Wu, Yu-Gang Jiang,
  and Ser-Nam Lim.
\newblock Adavit: Adaptive vision transformers for efficient image recognition.
\newblock In {\em Proceedings of the IEEE/CVF Conference on Computer Vision and
  Pattern Recognition}, pages 12309--12318, 2022.

\bibitem{misra2016cross}
Ishan Misra, Abhinav Shrivastava, Abhinav Gupta, and Martial Hebert.
\newblock Cross-stitch networks for multi-task learning.
\newblock In {\em Proceedings of the IEEE conference on computer vision and
  pattern recognition}, pages 3994--4003, 2016.

\bibitem{mohamed2021spatio}
Eslam Mohamed and Ahmad El~Sallab.
\newblock Spatio-temporal multi-task learning transformer for joint moving
  object detection and segmentation.
\newblock In {\em 2021 IEEE International Intelligent Transportation Systems
  Conference (ITSC)}, pages 1470--1475. IEEE, 2021.

\bibitem{neseem2021adacon}
Marina Neseem and Sherief Reda.
\newblock Adacon: Adaptive context-aware object detection for
  resource-constrained embedded devices.
\newblock In {\em 2021 IEEE/ACM International Conference On Computer Aided
  Design (ICCAD)}, pages 1--9. IEEE, 2021.

\bibitem{pytorch}
Pytorch mobile: End-to-end workflow from training to deployment for ios and
  android mobile devices, March 2023.
\newblock \url{https://pytorch.org/mobile/android}.

\bibitem{standley2020tasks}
Trevor Standley, Amir Zamir, Dawn Chen, Leonidas Guibas, Jitendra Malik, and
  Silvio Savarese.
\newblock Which tasks should be learned together in multi-task learning?
\newblock In {\em International Conference on Machine Learning}, pages
  9120--9132. PMLR, 2020.

\bibitem{strudel2021segmenter}
Robin Strudel, Ricardo Garcia, Ivan Laptev, and Cordelia Schmid.
\newblock Segmenter: Transformer for semantic segmentation.
\newblock In {\em Proceedings of the IEEE/CVF international conference on
  computer vision}, pages 7262--7272, 2021.

\bibitem{sun2020adashare}
Ximeng Sun, Rameswar Panda, Rogerio Feris, and Kate Saenko.
\newblock Adashare: Learning what to share for efficient deep multi-task
  learning.
\newblock {\em Advances in Neural Information Processing Systems},
  33:8728--8740, 2020.

\bibitem{deit}
Hugo Touvron, Matthieu Cord, Matthijs Douze, Francisco Massa, Alexandre
  Sablayrolles, and Herv{\'e} J{\'e}gou.
\newblock Training data-efficient image transformers \& distillation through
  attention.
\newblock In {\em International conference on machine learning}, pages
  10347--10357. PMLR, 2021.

\bibitem{mtl_survey}
Simon Vandenhende, Stamatios Georgoulis, Wouter Van~Gansbeke, Marc Proesmans,
  Dengxin Dai, and Luc Van~Gool.
\newblock Multi-task learning for dense prediction tasks: A survey.
\newblock {\em IEEE Transactions on Pattern Analysis and Machine Intelligence},
  44(7):3614--3633, 2022.

\bibitem{vandenhende2020mti}
Simon Vandenhende, Stamatios Georgoulis, and Luc Van~Gool.
\newblock Mti-net: Multi-scale task interaction networks for multi-task
  learning.
\newblock In {\em Computer Vision--ECCV 2020: 16th European Conference,
  Glasgow, UK, August 23--28, 2020, Proceedings, Part IV 16}, pages 527--543.
  Springer, 2020.

\bibitem{vaswani2017attention}
Ashish Vaswani, Noam Shazeer, Niki Parmar, Jakob Uszkoreit, Llion Jones,
  Aidan~N Gomez, {\L}ukasz Kaiser, and Illia Polosukhin.
\newblock Attention is all you need.
\newblock {\em Advances in neural information processing systems}, 30, 2017.

\bibitem{vuzix}
Vuzix m4000 smart glasses, March 2023.
\newblock \url{https://www.vuzix.com/products/m4000-smart-glasses}.

\bibitem{wang2018skipnet}
Xin Wang, Fisher Yu, Zi-Yi Dou, Trevor Darrell, and Joseph~E Gonzalez.
\newblock Skipnet: Learning dynamic routing in convolutional networks.
\newblock In {\em Proceedings of the European Conference on Computer Vision
  (ECCV)}, pages 409--424, 2018.

\bibitem{xu2018pad}
Dan Xu, Wanli Ouyang, Xiaogang Wang, and Nicu Sebe.
\newblock Pad-net: Multi-tasks guided prediction-and-distillation network for
  simultaneous depth estimation and scene parsing.
\newblock In {\em Proceedings of the IEEE Conference on Computer Vision and
  Pattern Recognition}, pages 675--684, 2018.

\bibitem{ye2022inverted}
Hanrong Ye and Dan Xu.
\newblock Inverted pyramid multi-task transformer for dense scene
  understanding.
\newblock In {\em Computer Vision--ECCV 2022: 17th European Conference, Tel
  Aviv, Israel, October 23--27, 2022, Proceedings, Part XXVII}, pages 514--530.
  Springer, 2022.

\bibitem{yu2022mia}
Zhongzhi Yu, Yonggan Fu, Sicheng Li, Chaojian Li, and Yingyan Lin.
\newblock Mia-former: efficient and robust vision transformers via
  multi-grained input-adaptation.
\newblock In {\em Proceedings of the AAAI Conference on Artificial
  Intelligence}, volume~36, pages 8962--8970, 2022.

\bibitem{zamir2018taskonomy}
Amir~R Zamir, Alexander Sax, William Shen, Leonidas~J Guibas, Jitendra Malik,
  and Silvio Savarese.
\newblock Taskonomy: Disentangling task transfer learning.
\newblock In {\em Proceedings of the IEEE conference on computer vision and
  pattern recognition}, pages 3712--3722, 2018.

\bibitem{bert_loses_patience}
Wangchunshu Zhou, Canwen Xu, Tao Ge, Julian McAuley, Ke Xu, and Furu Wei.
\newblock Bert loses patience: Fast and robust inference with early exit.
\newblock {\em Advances in Neural Information Processing Systems},
  33:18330--18341, 2020.

\end{thebibliography}
}

\end{document}